\documentclass[letterpaper,10pt,conference]{ieeeconf}

\IEEEoverridecommandlockouts
\usepackage{graphicx}
\usepackage{amsmath,amssymb}
\usepackage{booktabs}
\usepackage{multirow}
\usepackage{array}
\usepackage{xcolor}
\usepackage[normalem]{ulem}
\usepackage{cite}
\usepackage{url}
\usepackage{balance}
\usepackage{float}
\usepackage[caption=false,font=footnotesize]{subfig}

\title{\LARGE \bf

From Prediction to Action:
Learning Physical Context \\for Deformable Linear Object Manipulation

}

\author{Chengxiao Dong, Alessio Caporali, Hongyu Lan and Gianluca Palli%
\thanks{The authors are with DEI - Department of Electrical, Electronic and Information Engineering, University of Bologna, 40136 Bologna, Italy.}%
\thanks{Corresponding author: hongyu.lan@unibo.it}%
\thanks{This work was supported by the China Scholarship Council (CSC).}%
\thanks{The authors acknowledge Algoryx Simulation AB for providing an academic license of AGX Dynamics, which enabled the high-fidelity simulation studies conducted in this work.}%
}

\begin{document}
\maketitle
\thispagestyle{empty}
\pagestyle{empty}

\begin{abstract}

A representation that accurately predicts a deformable object's response is not necessarily the one most useful for choosing manipulation actions. We study this distinction between \emph{predictive utility} and \emph{decision utility} in planar single-grasp shape control of Deformable Linear Objects (DLOs). As a common decision backbone, we introduce Geometry--Pointer--Flow (GPF), a geometry-grounded policy that separates grasp-location selection from grasp-conditioned motion generation.
Its geometry-only variant reduces one-step executed shape error by 32.8\% relative to a DLO-specific diffusion policy, while using 18.7$\times$ fewer parameters and achieving approximately 193$\times$ faster inference.
On this backbone, we compare physical context represented by ground-truth parameters, estimated parameters, and learned interaction-history representations, for both forward prediction and executed control. 
For prediction, ground-truth physical parameters reduce next-shape error by 15.8\% over geometry alone, and estimated parameters are slightly more predictive than interaction history. This ordering does not carry
over to control: estimated parameters provide no detectable one-step control gain, whereas history-conditioned GPF reduces executed error by 9.9\% relative to geometry-only GPF across 470 paired simulation trials. In closed loop, interaction history improves strict-threshold attainment in simulation and yields the strongest zero-shot performance on hardware.
All policies are trained exclusively in simulation and transferred to the physical robot without real-world training or fine-tuning. 
These results show that physical context should be evaluated by the decisions it supports rather than by predictive accuracy alone. Code will be released upon acceptance.

\end{abstract}


\section{Introduction}

Deformable linear objects (DLOs), such as cables, ropes, hoses, and wires, are challenging to manipulate because similar geometries can respond differently to the same action due to variations in stiffness, friction, and interaction history \cite{caporali2026robotic}. Effective manipulation therefore requires reasoning about both geometry and physical response.

This has motivated two complementary directions: learning DLO dynamics models to predict action outcomes
\cite{yang2022dlodynamics,zhang2025particlegrid,ha2025inrdom}, and
conditioning manipulation policies on explicit physical parameters or
learned latent representations \cite{kuroki2024gendom,li2024flexibility}.
More broadly, online system identification and rapid adaptation exploit similar information to adapt control under changing physical conditions \cite{yu2017uposi,kumar2021rma,liang2024rma_manipulation}. 
Throughout this work, we use the term \emph{physical context} to denote information that characterizes how the object responds to interaction, whether through explicit parameters or learned latent representations.

This raises a fundamental question about the role of physical context: \emph{is the information that best predicts an object's response also the information most useful for choosing how to manipulate it?} The two objectives may not align: a representation may be highly informative for next-state prediction yet less useful for action selection if it does not preserve control-relevant information. This distinction is closely related to control-oriented system identification, where improved predictive accuracy does not necessarily yield improved control performance \cite{sivaranjani2026controloriented}.
For learned policies, an additional challenge is how physical context is represented and integrated into the policy.

\begin{figure*}[t]
    \centering
    \includegraphics[width=\textwidth]{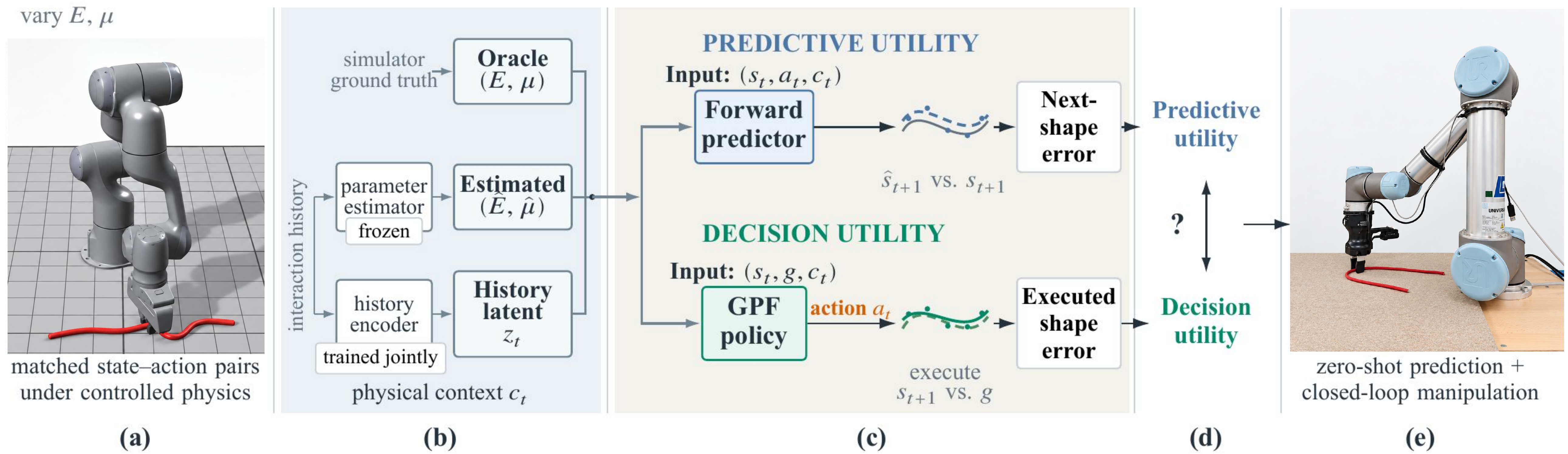}
    \caption{
    Overview of the \emph{predictive} and \emph{decision utility} of \emph{physical context}.
    (a) Controlled simulation generates matched state--action pairs while varying stiffness $E$ and friction $\mu$.
    (b) Three alternative physical contexts $c_t$, one of which conditions the models in (c): Oracle uses the simulator's ground-truth parameters $(E,\mu)$; Estimated infers $(\hat E,\hat\mu)$ from the interaction history with a pretrained, frozen parameter estimator; Hist encodes the same history into a latent $z_t$ with an encoder trained jointly with the downstream model. None denotes the geometry-only baseline.
    (c) The same context $c_t$ conditions either a forward predictor, whose next-shape error defines \emph{predictive utility}, or a goal-conditioned GPF policy, whose executed shape error defines \emph{decision utility}.
    (d) We test whether predictive utility aligns with decision utility.
    (e) Deployable context representations are evaluated zero-shot on the real robot for next-shape prediction and closed-loop manipulation.
    }
    \label{fig:overview}
\end{figure*}

In this work, we distinguish between the \emph{predictive utility} of
physical context, namely its ability to improve prediction of an action's outcome, and its \emph{decision utility}, namely its ability to improve action selection. 
We study this distinction in planar single-grasp DLO shape manipulation by comparing three physical-context representations (\emph{Oracle}, \emph{Estimated}, and \emph{Hist}) against a geometry-only baseline.
Unlike the explicit parameterizations, interaction-history context is not constrained to recover predefined physical quantities and can retain information from recent action--response history. Predictive utility is evaluated with a separate forward-response model, while decision utility is evaluated through a learned manipulation policy. Fig.~\ref{fig:overview} summarizes the study: controlled simulation (a) generates alternative physical contexts (b), whose predictive and decision utility (c) are compared (d) before real-world zero-shot evaluation (e).

We study this question primarily in simulation using AGX Dynamics
\cite{Algoryx_AGX_Dynamics,Lacoursiere_2007}, which provides controlled physics-based interactions and known material and contact parameters for comparing physical context in prediction and control. To assess decision utility without imposing a predefined physical model, we condition a common learned policy on each context and evaluate executed control. The resulting framework compares explicit parameters and interaction representations by how well they support action selection. We instantiate it with Geometry--Pointer--Flow (GPF), a geometry-grounded policy that separates \emph{where} to grasp the DLO from \emph{how} to move the selected grasp-point.

Our evaluation first establishes GPF as a strong manipulation backbone, then compares the predictive and decision utility of physical context. Context that improves prediction does not necessarily benefit action selection: Estimated parameters substantially improve forward prediction but provide no detectable executed-control gain, whereas interaction-derived history context gives the strongest observed control performance among deployable representations. 
We finally evaluate zero-shot transfer from simulation to real-world DLO manipulation, without real-world policy training or fine-tuning.

To summarize, the main contributions of this work are:

\begin{itemize}

    \item \textbf{Predictive versus decision utility of physical context.}
    We compare explicit physical parameters, estimated parameters, and interaction-history context across prediction and control, showing that improved response prediction does not necessarily translate into improved action selection.

    \item \textbf{A geometry-grounded policy for DLO manipulation.}
    We introduce GPF, which separates grasp-location selection from
    motion generation and provides a common backbone for
    evaluating the decision utility of alternative physical contexts.

    \item \textbf{Interaction-history adaptation and zero-shot validation.}
    We show that history-derived context improves strict-threshold attainment in simulation and transfers zero-shot to the physical robot, where it yields the strongest closed-loop performance among the deployed controllers.

\end{itemize}

\section{Related Work}
\label{sec:related_works}

\subsection{DLO Manipulation and Dynamics Learning}

Learning-based DLO manipulation includes direct policy learning and
model-based control. Structured methods couple contact selection with motion \cite{wu2020deformable,seita2021goalconditioned}. A complementary line learns DLO dynamics using locally linear latent models \cite{zhang2021latentdlo}, differentiable state-space models
\cite{yang2022dlodynamics}, graph-based models
\cite{wang2022offlineonline}, and richer visual representations
\cite{zhang2025particlegrid,ha2025inrdom}. These methods show that learned dynamics support prediction and control, but do not test whether the same physical context is equally useful for both.

\subsection{Structured and Generative Manipulation Policies}

GPF relates to discrete--continuous policies that separate interaction location from motion. HACMan conditions continuous motion on a selected contact \cite{zhou2023hacman}, with related hybrid formulations in RL and robotic assembly \cite{li2022hyar,zhang2021insertion}. Generative policies
include Diffusion Policy \cite{chi2025diffusionpolicy}, DLO-specific action diffusion \cite{mengozzi2025actiondiffusion}, and robot policies built on flow matching and rectified flow \cite{lipman2023flowmatching,liu2023rectifiedflow,
braun2024rfmp,chisari2024pointflowmatch}. GPF combines geometry-grounded contact selection with flow-based motion generation and serves as a shared backbone for comparing physical contexts under one action-selection structure.

\subsection{Physical Context for Adaptation}

Adaptive policies condition control on explicit parameters or learned latent context. Online system identification estimates environment parameters \cite{yu2017uposi}, Projected Universal Policies learn low-dimensional control-oriented representations \cite{yu2019pup}, and RMA infers latent context from recent interactions \cite{kumar2021rma, liang2024rma_manipulation}. For deformable objects, prior work estimates physical parameters for adaptation and optimization \cite{caporali2024deformable}, conditions policies on estimated properties \cite{kuroki2024gendom}, and uses estimated flexibility for DLO insertion \cite{li2024flexibility}. We instead evaluate physical context separately for forward-response prediction and executed manipulation, testing whether predictive utility transfers to decision utility.

\section{Methodology}
\label{sec:method}

\subsection{Problem Formulation}

We consider planar single-grasp manipulation of a DLO. At interaction $t$, the DLO centerline is represented by $n$ ordered points $s_t=(p_{t,1},\ldots,p_{t,n})$, with $p_{t,i}\in\mathbb{R}^2$, and the desired shape by $g=(p^g_1,\ldots,p^g_n)$.

\begin{figure}[!t]
    \centering
    \includegraphics[width=\columnwidth]{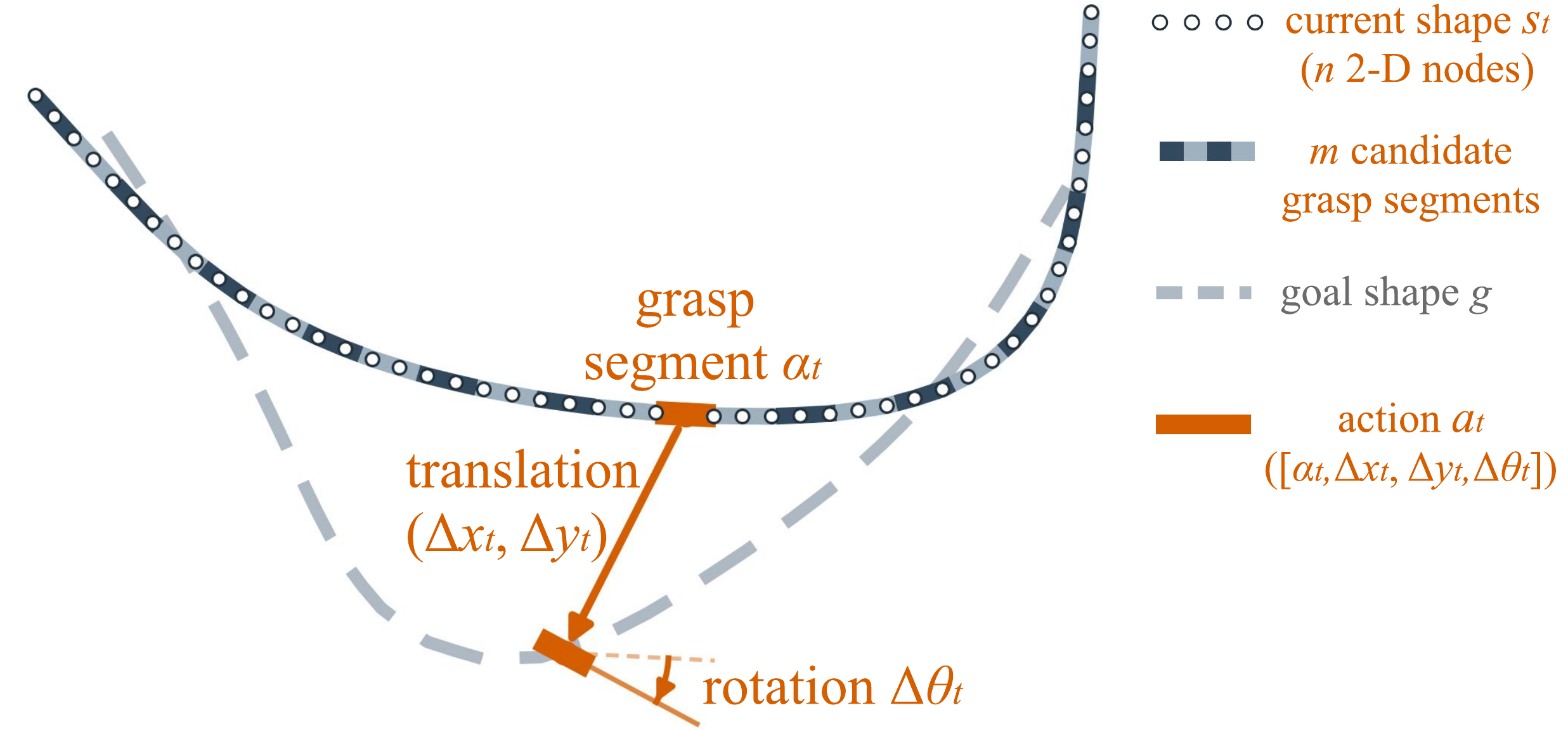}
    \caption{
    State and hybrid-action representation for planar single-grasp DLO manipulation. The current shape $s_t$ and goal $g$ are represented by $n$ two-dimensional centerline nodes. The action selects one of $m$ candidate grasp segments and specifies its planar translation and rotation.
    }
    \label{fig:hybrid_action}
\end{figure}

As shown in Fig.~\ref{fig:hybrid_action}, each action
selects one of $m$ candidate grasp segments and specifies a planar
motion. We write
\begin{equation}
    a_t = [\alpha_t,\Delta x_t,\Delta y_t,\Delta\theta_t],
    \qquad
    \alpha_t \in \{1,\ldots,m\},
\end{equation}
where $\alpha_t$ denotes the selected grasp segment and
$(\Delta x_t,\Delta y_t,\Delta\theta_t)$ the corresponding translation and rotation.

After executing $a_t$, the robot observes the resulting centerline $s_{t+1}$; the objective is to select actions that drive the centerline toward the target $g$. Geometry and actions are expressed
in a normalized, object-aligned representation. Discretization, action bounds, and normalization constants are given in Sec.~\ref{sec:exp_dataset_simulation}.

\subsection{Physical-Context Representations}

As outlined in Fig.~\ref{fig:overview}(b), 
we consider three physical-context representations $c_t$---Oracle,
Estimated, and Hist---together with a geometry-only baseline. \emph{None} uses geometry alone. \emph{Oracle} provides the simulator's Young's modulus $E$ and DLO--surface friction coefficient $\mu$. \emph{Estimated} uses estimates $(\hat E_t,\hat \mu_t)$ of the same quantities inferred from previous interactions. \emph{Hist} maps the same interaction history directly to a learned latent $z_t$, without constraining it to recover predefined physical parameters.

For both Estimated and Hist, the input consists of up to five completed interactions, each including the pre-action DLO centerline, selected grasp, executed motion, and resulting centerline deformation. Only interactions completed before the current action are used, so the current response is never available to the context. The two representations therefore differ only in how the same history is compressed (Fig.~\ref{fig:overview}(b)): Estimated imposes the explicit physical bottleneck $(E,\mu)$, whereas Hist learns a task-dependent latent representation.

\subsubsection{Explicit Physical Parameters}

Estimated infers $(\hat E_t,\hat\mu_t)$ from interaction quantities expressed in a grasp-centered tangent--normal frame, which reduces dependence on absolute object pose.

Separate residual point encoders extract features for $E$ and $\mu$,
followed by independent two-layer temporal Transformers over the
interaction sequence. The estimator is trained using weighted Huber
regression together with a small cross-target decorrelation penalty and is frozen when used downstream. Without previous interactions, it returns the training-set prior.

\subsubsection{Latent Interaction-History Context}

Hist removes the explicit $(E,\mu)$ bottleneck and instead encodes the same interaction history into a 32-dimensional latent representation $z_t$.

For each interaction, node-wise position, deformation, arc length, and distance from the selected grasp are used to preserve the spatial
structure of the DLO response. One-dimensional convolutional blocks
encode individual interactions, followed by temporal attention over the interaction sequence to produce \(z_t\). Unlike Estimated, the encoder is trained jointly with the downstream predictor or policy and can retain task-relevant information.

An auxiliary head predicts normalized $E$ and $\mu$ from $z_t$ with a squared-error loss $\mathcal{L}_{\mathrm{aux}}$, used only during training, to encourage physically meaningful representations. Missing interactions are masked, while empty-history samples allow the model to operate before any interaction has been observed.

\subsection{Geometry--Pointer--Flow Manipulation Policy}
\label{sec:gpf}

\begin{figure*}[!t]
    \centering
    \includegraphics[width=\textwidth]{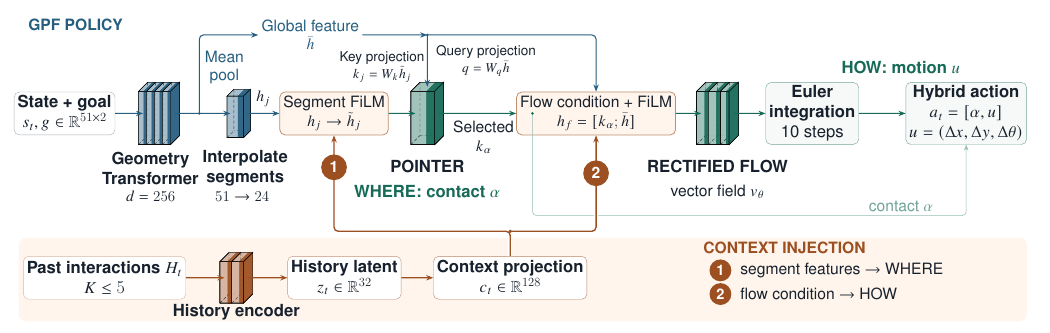}
    \caption{
    Geometry--Pointer--Flow policy with interaction-history conditioning.
    GPF selects \emph{where} to interact through a query--key pointer and generates \emph{how} to move the selected contact through rectified flow.
    The latent context modulates segment features before grasp selection (1) and the post-selection flow condition (2).
    }
    \label{fig:gpf}
\end{figure*}

GPF is the common policy backbone used to evaluate \emph{decision utility} (Fig.~\ref{fig:overview}(c)). 
Given the current shape \(s_t\), target shape \(g\), and optional physical context \(c_t\), GPF predicts the hybrid action \(a_t=\pi(s_t,g,c_t)\) by separating \emph{where} to interact with the DLO from \emph{how} to move the selected contact (Fig.~\ref{fig:gpf}).

A shared geometry Transformer encodes the current and target centerlines from seven-dimensional node features (current position, target position, displacement, and arc length). Node features are interpolated at the candidate grasp segments and shared by both stages: a pointer selects the grasp, and a conditional rectified-flow model generates the planar motion.

\subsubsection{Grasp Selection}

Let \(h^{\mathrm{node}}_{t,i}\in\mathbb{R}^d\) denote the encoded feature of node \(i\), and \(h_{t,j}\in\mathbb{R}^d\) the feature interpolated at candidate grasp segment \(j\). Mean pooling over the node features provides a global geometry representation \(\bar h_t\).

For geometry-only GPF, the global representation acts as a query and
each candidate segment as a key. The resulting grasp scores are
\begin{equation}
    \ell_{t,j}
    =
    \frac{(W_q\bar h_t)^\top(W_kh_{t,j})}{\sqrt{d}},
    \qquad
    \hat{\alpha}_t = \arg\max_j \ell_{t,j}.
\end{equation}
The demonstrated grasp segment \(\alpha_t\) supervises the pointer
through categorical cross-entropy, denoted by
\(\mathcal{L}_{\mathrm{ptr}}\).

\subsubsection{Motion Generation}

Given a selected grasp segment \(\alpha\), GPF generates the continuous motion \(u_t=(\Delta x_t,\Delta y_t,\Delta\theta_t)\) conditioned on both the global DLO geometry and the local representation of the selected contact. The base flow condition is
\(h_{f,t}=[k_{t,\alpha};\bar h_t]\), where
\(k_{t,\alpha}=W_kh_{t,\alpha}\). This couples motion generation to the chosen location. A rectified-flow model conditioned on \(h_{f,t}\) generates \(u_t\): let \(x_0\sim\mathcal{N}(0,I)\), \(x_1=u_t\), and
\(\tau\sim\mathcal{U}[0,1]\), with
\(x_\tau=(1-\tau)x_0+\tau x_1\). The flow objective is
\begin{equation}
    \mathcal{L}_{\mathrm{flow}}
    =
    \left\|
    v_\theta(x_\tau,\tau\mid h_{f,t})-(x_1-x_0)
    \right\|_2^2.
\end{equation}

\subsubsection{Physical-Context Conditioning}

Physical context is injected only after geometry encoding, so all policy variants use the same geometry-encoder architecture.
For the concatenation-based variants, Oracle, Estimated, and Hist
contexts are projected to a 128-dimensional embedding
$\tilde c_t$.

Context affects both stages of GPF. For grasp selection, it augments the global query while the segment keys remain geometry-based,
\begin{equation}
    q_t = W_q[\bar h_t;\tilde c_t],
    \qquad
    k_{t,j}=W_kh_{t,j}.
\end{equation}
For motion generation, the same context is appended to the flow
condition,
\begin{equation}
    h_{f,t} = [k_{t,\alpha};\bar h_t;\tilde c_t].
\end{equation}
Oracle uses the ground-truth \((E,\mu)\), Estimated uses
\((\hat E_t,\hat\mu_t)\) from the frozen parameter estimator, and
Hist-concat uses the interaction-history latent \(z_t\).

We additionally consider Hist-FiLM, where \(z_t\) is projected to
\(c_t\) and injected through feature-wise linear modulation (FiLM)
\cite{perez2018film},
\begin{equation}
    \mathrm{FiLM}(h,\tilde c_t)
=
h\odot\bigl(1+\gamma(\tilde c_t)\bigr)+\beta(\tilde c_t),
\end{equation}
with zero-initialized \(\gamma(\cdot)\) and \(\beta(\cdot)\).
FiLM modulates the candidate segment features before grasp selection and the flow condition after it (Fig.~\ref{fig:gpf}), so physical context affects both \emph{where} and \emph{how} the DLO is manipulated.

Hist-concat and Hist-FiLM share the history formulation but are trained end-to-end with separate encoders, so their comparison evaluates complete conditioning strategies rather than the fusion operator alone.

\subsubsection{Training and Inference}

Training proceeds through grasp-selection, motion-generation, and joint stages. The motion-generation stage conditions on the demonstrated grasp $\alpha_t$; joint training uses it with probability \(0.75\) and the pointer prediction \(\hat{\alpha}_t\) otherwise. At inference, motion generation always uses the predicted grasp.

The joint objective is
\begin{equation}
    \mathcal{L}_{\mathrm{joint}}
    =
    \mathcal{L}_{\mathrm{ptr}}
    +
    \mathcal{L}_{\mathrm{flow}}
    +
    \lambda_{\mathrm{aux}}\mathcal{L}_{\mathrm{aux}},
\end{equation}
where \(\lambda_{\mathrm{aux}}=3\) when the Hist encoder is trained
jointly with the policy and \(0\) otherwise. During the dedicated motion-generation stage, the Hist encoder is frozen and the auxiliary term is inactive.
At inference, the vector field is integrated from Gaussian noise using 10 Euler steps. Here \(\tau\) denotes flow time and is distinct from the interaction index \(t\).

\subsection{Forward-Response Prediction}
\label{sec:forward}

To evaluate \emph{predictive utility} independently of action selection (Fig.~\ref{fig:overview}(c)), we train a forward model to predict the DLO deformation produced by an executed action,

\begin{equation}
    F_\phi(s_t,a_t,c_t)\rightarrow\widehat{\Delta s}_{t+1},
\qquad
\hat{s}_{t+1}=s_t+\widehat{\Delta s}_{t+1}.
\end{equation}

Unlike the manipulation policy, the predictor receives the executed
action \(a_t\), does not receive the target shape \(g\), and never
selects an action.

A Transformer encodes the current DLO geometry; the executed action is combined with the global and selected-segment representations before per-node displacement decoding, and the physical context is projected and injected into the action representation through zero-initialized FiLM.

The predictor is trained using the mean squared node-displacement error,
\begin{equation}
    \mathcal{L}_{\mathrm{pred}}
    =
    \frac{1}{n}
    \sum_{i=1}^{n}
    \left\|
    \widehat{\Delta p}_{t+1,i}
    -
    \Delta p_{t+1,i}
    \right\|_2^2,
\end{equation}
where
\(\Delta p_{t+1,i}=p_{t+1,i}-p_{t,i}\).

\emph{None}, \emph{Oracle}, \emph{Estimated}, and \emph{Hist} are trained as independent predictors with the same conditioning architecture and injection point, sharing no weights. For Hist, the interaction-history encoder is
trained jointly with the predictor using
\begin{equation}
    \mathcal{L}_{\mathrm{pred}}^{\mathrm{Hist}}
    =
    \mathcal{L}_{\mathrm{pred}}
    +
    3\mathcal{L}_{\mathrm{aux}}.
\end{equation}

\section{Experimental Setup}
\label{sec:exp_setup}

\subsection{Dataset and Simulation}
\label{sec:exp_dataset_simulation}

AGX Dynamics \cite{Algoryx_AGX_Dynamics} is used for all training and
simulation-based evaluation (Fig.~\ref{fig:overview}(a)).
The DLO is modeled as a 0.5\,m cable, discretized into 24 constrained
segments and resampled as a 51-point planar centerline.
We vary the DLO Young's modulus as
$E\in[20,200]\,\mathrm{kPa}$, sampled log-uniformly, and the
DLO--surface friction coefficient as
$\mu\in[0.15,0.60]$, sampled uniformly.

The dataset contains 90,106 episodes, corresponding to 901,060
transitions. Episodes are partitioned into training, validation, and test sets using an 80/10/10 split over both episodes and physical conditions.

States and actions follow the representation in
Fig.~\ref{fig:hybrid_action}. Geometry is PCA-aligned, with planar
translation bounded to $[-0.2,0.2]^2\,\mathrm{m}$ and normalized by
$0.2\sqrt{2}\,\mathrm{m}$, while rotation is normalized by $\pi/4$.

Each episode uses fixed $(E,\mu)$ and starts from a randomly oriented straight DLO that is allowed to settle; feasible actions are then sampled from the settled configuration, and the resulting next states serve as physically reachable one-step targets.

\subsection{Manipulation Policy Baselines}\label{sec:baselines}

We compare GPF against behavioral-cloning and generative-policy baselines.
\emph{MLP-BC} regresses the hybrid action from flattened current--target geometry, while \emph{Transformer-BC} uses the GPF geometry encoder followed by global pooling and action regression.
\emph{Global-Flow} uses conditional flow matching on the global geometry representation, without conditioning motion on a selected grasp.
\emph{GPF-Geom} adds the proposed where/how factorization, with pointer-based grasp selection and motion generation conditioned on the selected segment.
\emph{DLO-DP} re-implements the DLO-specific diffusion policy following
\cite{mengozzi2025actiondiffusion}: a 100-step DDPM generates a
$16\times5$ action trajectory from global current--target geometry, and its continuous grasp coordinate is mapped to one of 24 candidate segments.
All methods use identical data splits and update budgets, with checkpoint selection based on validation performance.

\subsection{Evaluation Protocol}

Executed control is evaluated using the mean-node shape error,
\begin{equation}
    e_{\mathrm{shape}}
    =
    \frac{1}{n}
    \sum_{i=1}^{n}
    \left\|
    p_{t+1,i}-p_i^{g}
    \right\|_2,
\end{equation}
reported in millimeters with consistent node ordering.
Batch-1 inference latency is measured over 100 synchronized FP32 runs on an RTX 2080 Ti after 20 warm-up runs. Unless stated otherwise, paired 95\% confidence intervals are computed using a 5,000-replicate percentile bootstrap; an effect is considered detectable when the interval excludes zero.

Predictive utility is evaluated on 8,192 held-out transitions using
independently trained None, Oracle, Estimated, and Hist forward predictors.
Decision utility is evaluated on 470 paired one-step AGX trials using
GPF-Geom, Oracle-concat, Estimated-concat, Hist-concat, and Hist-FiLM. One
transition is sampled from each of 470 distinct test episodes. For each trial, the recorded interaction prefix is replayed deterministically with zero discrepancy, after which all methods are executed from the same state, target, physical parameters, and history; no executions failed. Context-conditioned DLO-DP variants and the policy baselines of Sec.~\ref{sec:baselines} are evaluated using the same protocol.

Closed-loop simulation compares DLO-DP, GPF-Geom, and Hist-FiLM over 100 rollouts per controller, each with a 10-action horizon. Target attainment is measured at thresholds of $10$, $8$, and $5\,\mathrm{mm}$. Hist-FiLM starts each rollout with empty history and builds its context only from interactions within that rollout.

\begin{figure}[t]
    \centering
    \includegraphics[width=0.8\columnwidth]{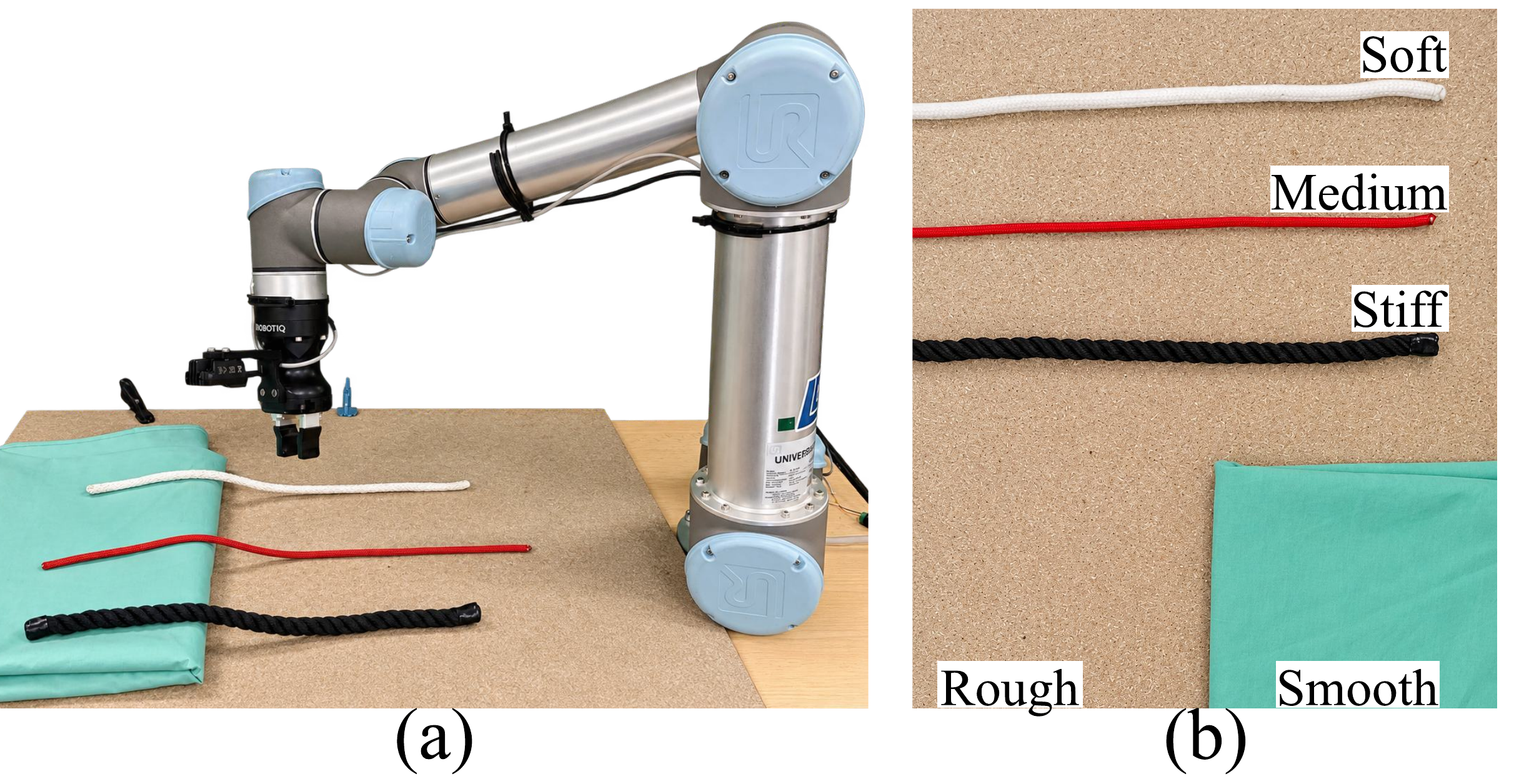}
    \caption{
    Zero-shot real-robot evaluation.
    (a) UR5 platform with a Robotiq Hand-E gripper and wrist-mounted camera.
    (b) Three DLO compliance levels and the two test surfaces.
    }
    \label{fig:real}
    \vspace{-4mm}
\end{figure}

\subsection{Real-Robot Evaluation}
\label{sec:real_robot}

We evaluate zero-shot sim-to-real transfer on a UR5 equipped with a
Robotiq Hand-E gripper and a calibrated wrist-mounted OAK camera
(Fig.~\ref{fig:real}(a)). All models are trained exclusively in AGX
Dynamics, where actions are executed by a simulated Franka Emika Panda,
and transferred to the physical setup without real-world training or
fine-tuning (Fig.~\ref{fig:overview}(e)). Because the hybrid action is defined relative to the DLO rather than in robot coordinates, changing manipulator and gripper requires no policy modification.

\begin{figure}[!t]
    \centering
    \includegraphics[width=0.9\columnwidth]
    {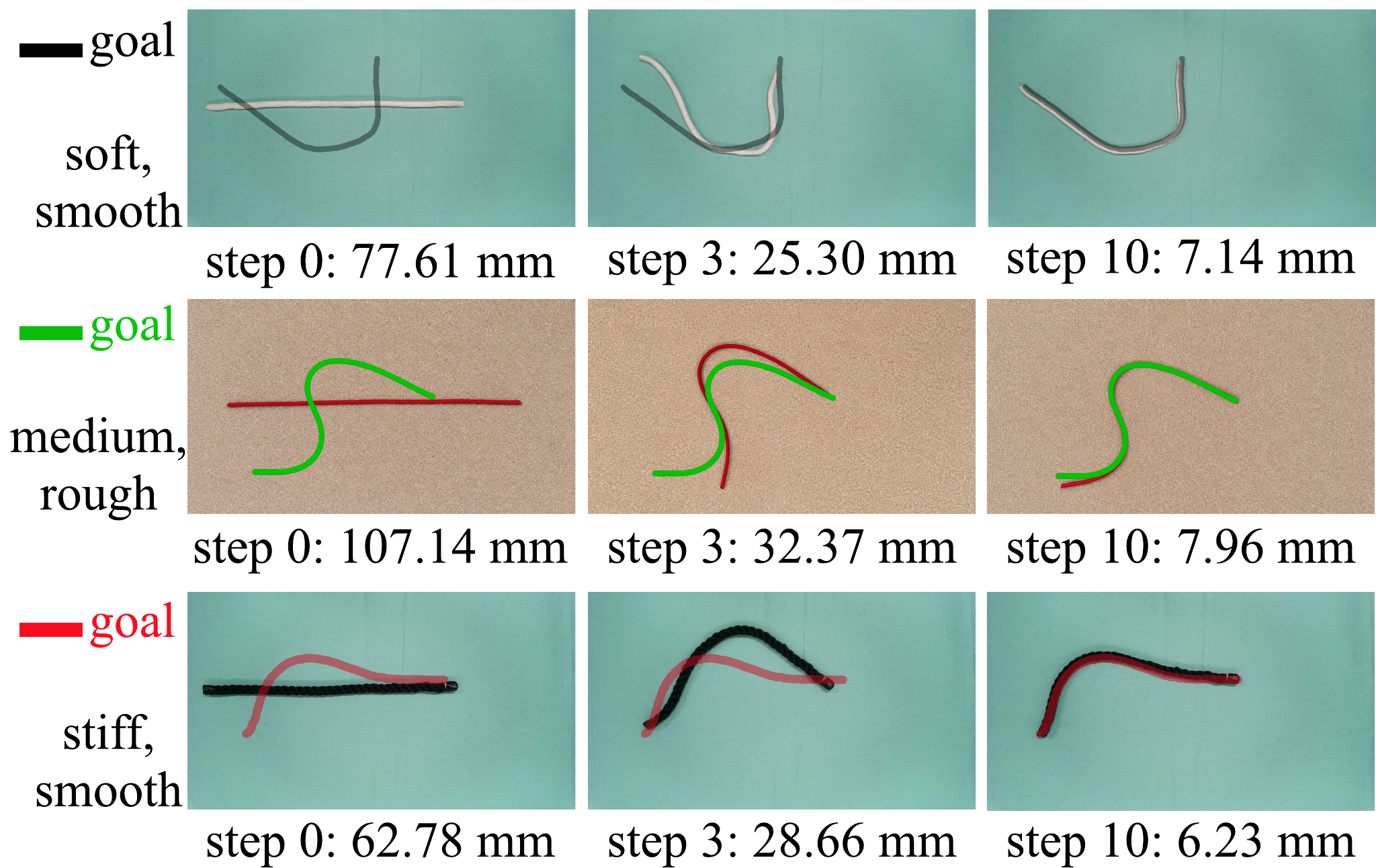}
    \caption{
    Representative zero-shot real-robot manipulation processes of GPF Hist-FiLM under three
    DLO--surface conditions. Columns show steps 0, 3, and 10.
    Colored curves denote target shapes, and values report mean-node shape error.
    }
    \label{fig:realrobot_process}
    \vspace{-3mm}
\end{figure}

Three DLO compliance levels (soft, medium, and stiff) and two support
surfaces (smooth and rough) define six test conditions
(Fig.~\ref{fig:real}(b)). The rough surface is bare cork; the smooth surface covers the same board with a smooth cloth.
The DLO centerline is extracted from the camera image following
\cite{caporali2022fastdlo} and resampled to the same 51-node centerline
and 24 candidate grasp segments used in simulation.
After each action, the robot releases the DLO and returns to a fixed observation pose, from which the wrist-mounted camera observes the whole DLO before the next action is selected.
Representative zero-shot closed-loop manipulation processes are shown in
Fig.~\ref{fig:realrobot_process}.

Each condition contains 10 target shapes. DLO-DP, GPF-Geom, and Hist-FiLM
are each evaluated over 60 complete 10-action rollouts, yielding 1,800
action transitions in total. Controller order is randomized within
target--condition blocks, while manual reset prevents exact replay of the
same physical initial state.

Real-world prediction is evaluated on all 1,800 transitions. Matched context
is taken from the query rollout, whereas each mismatch category pairs every
query exhaustively with all eligible samples from the corresponding alternative
condition. Closed-loop performance is evaluated post hoc using rollout-best
error, 10\,mm target attainment, and first passage to 10\,mm among successful
trials; passage time is interpreted jointly with attainment.

\section{Results} \label{sec:results}

We first evaluate the GPF architecture (Sec.~\ref{sec:results_gpf}), then compare the predictive and decision utility of physical context (Secs.~\ref{sec:results_prediction} and \ref{sec:results_decision}), and finally evaluate closed-loop behavior in simulation and on the real robot (Sec.~\ref{sec:results_closed_loop}).

\subsection{GPF Architecture Evaluation}
\label{sec:results_gpf}

Table~\ref{tab:gpf} compares the policy architectures under paired execution.
\emph{GPF-Geom} achieves the lowest executed error while using 18.7$\times$ fewer parameters and substantially lower latency than \emph{DLO-DP}. All paired difference intervals exclude zero. \emph{Global-Flow} performs similarly to \emph{DLO-DP}, suggesting that flow matching alone does not account for the improvement. The result is instead consistent with the proposed \emph{where}/\emph{how} factorization, which combines explicit grasp
selection with motion generation conditioned on the selected segment and its local geometry.

\begin{table}[t]
    \centering
    \caption{\centering
    Policy architecture comparison on 470 paired simulation AGX trials.
    $\Delta$ error is relative to GPF-Geom (positive values indicate worse performance). CIs use paired percentile bootstrap.
    }
    \label{tab:gpf}
    \scriptsize
    \setlength{\tabcolsep}{2.0pt}
    \resizebox{0.8\columnwidth}{!}{%
    \begin{tabular}{@{}lccccc@{}}
        \toprule
        Method
        & Error
        & $\Delta$
        & 95\% CI
        & Params
        & Lat. \\
        & (mm)
        & (mm)
        & (mm)
        & (M)
        & (ms) \\
        \midrule

        DLO-DP
        & 13.782
        & $+4.525$
        & $[4.136,\,4.895]$
        & 75.66
        & 851.04 \\

        MLP-BC
        & 14.317
        & $+5.060$
        & $[4.733,\,5.378]$
        & 3.88
        & 0.32 \\

        Transformer-BC
        & 13.957
        & $+4.700$
        & $[4.476,\,4.950]$
        & 3.76
        & 1.37 \\

        Global-Flow
        & 13.485
        & $+4.228$
        & $[4.050,\,4.375]$
        & 3.79
        & 3.88 \\

        \textbf{GPF-Geom}
        & \textbf{9.257}
        & --
        & --
        & 4.05
        & 4.41 \\

        \bottomrule
    \end{tabular}
    }
    \vspace{-3mm}
\end{table}

\begin{figure}[!t]
    \centering
    \makebox[\columnwidth][c]{%
        \subfloat[]{%
            \includegraphics[height=0.37\columnwidth]
            {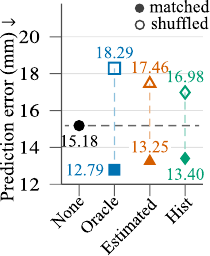}
            \label{fig:context_utility_a}
        }
        \hspace{1mm}
        \subfloat[]{%
            \includegraphics[height=0.37\columnwidth]
            {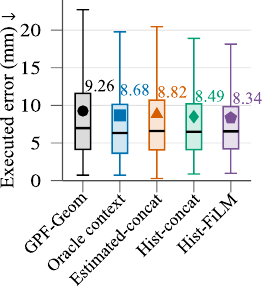}
            \label{fig:context_utility_b}
        }
        \hspace{1mm}
        \subfloat[]{%
            \includegraphics[height=0.37\columnwidth]
            {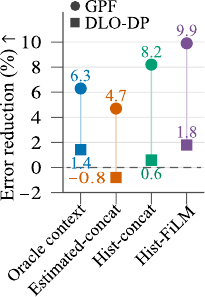}
            \label{fig:context_utility_c}
        }
    }

    \caption{
    Simulation evaluation of the \emph{predictive} and \emph{decision}
    utility of physical context.
    (a) Next-shape prediction error on held-out test transitions; filled
    and open markers indicate matched and shuffled context, respectively.
    (b) Executed one-step error over 470 paired AGX trials. Boxes show
    Q1--Q3 and the median, markers show means, and whiskers extend to
    $1.5\times$IQR; outliers are omitted. The black circle denotes the
    None baseline.
    (c) Executed-error reduction from context conditioning for GPF and DLO-DP.
    }
    \label{fig:context_utility}
    \vspace{-3mm}
\end{figure}

\subsection{Predictive Utility of Physical Context}
\label{sec:results_prediction}

\subsubsection{Simulation}

After five interactions, the explicit estimator achieves mean absolute errors of 17.37\,kPa for $E$ and 0.0700 for $\mu$. Physical context improves forward prediction
(see Fig.~\ref{fig:context_utility}(a)): \emph{Oracle} outperforms \emph{None} by 2.391\,mm
(95\% CI $[2.290,2.489]$), while \emph{Estimated} outperforms
\emph{Hist} by 0.146\,mm (95\% CI $[0.095,0.198]$).
Shuffling context while keeping predictor weights fixed degrades all
variants; for \emph{Hist}, the error increases by 3.58\,mm
(95\% CI $[3.422,3.754]$). Thus, the prediction gains depend on matched,
sample-specific context rather than model capacity alone.

\subsubsection{Real robot}

Prediction is evaluated on all 1,800 real-robot transitions
(Sec.~\ref{sec:real_robot}); unlike the simulation trials, physical initial
states are not exactly replayed across controllers.
The simulation ranking does not transfer unchanged to hardware
(Table~\ref{tab:real_prediction_context}). Matched Hist achieves the lowest overall error, whereas matched Estimated performs worse than None.
For Hist, mismatching the DLO produces the largest degradation.
Differences among mismatch categories are descriptive, as paired confidence intervals are unavailable. Overall, predictive utility depends on both the representation and its sim-to-real transfer.

\begin{table}[!t]
    \centering
    \caption{
    Zero-shot real-robot next-shape prediction error (mm) over 1,800 transitions. Context is matched or mismatched by rollout, DLO, or surface; bold indicates the row minimum.
    }
    \label{tab:real_prediction_context}
    \scriptsize
    \setlength{\tabcolsep}{1.5pt}
    \renewcommand{\arraystretch}{1.05}
    \resizebox{\columnwidth}{!}{%
    \begin{tabular}{llc cc cc cc cc}
        \toprule
        & &
        \multicolumn{1}{c}{Baseline} &
        \multicolumn{2}{c}{Matched} &
        \multicolumn{2}{c}{Same cond.} &
        \multicolumn{2}{c}{Other DLO} &
        \multicolumn{2}{c}{Other surface} \\
        \cmidrule(lr){3-3}
        \cmidrule(lr){4-5}
        \cmidrule(lr){6-7}
        \cmidrule(lr){8-9}
        \cmidrule(lr){10-11}

        DLO & Surface &
        None &
        Est. & Hist. &
        Est. & Hist. &
        Est. & Hist. &
        Est. & Hist. \\
        \midrule

        Soft & Smooth &
        8.91 &
        9.90 & \textbf{8.24} &
        10.03 & 8.34 &
        10.10 & 8.97 &
        10.02 & 8.60 \\

        Soft & Rough &
        8.72 &
        9.32 & 7.76 &
        9.24 & \textbf{7.72} &
        9.32 & 9.23 &
        9.24 & 8.14 \\

        Medium & Smooth &
        10.29 &
        11.77 & \textbf{8.62} &
        11.80 & 9.14 &
        11.83 & 9.70 &
        10.19 & 8.73 \\

        Medium & Rough &
        10.84 &
        13.91 & \textbf{9.44} &
        13.88 & 9.66 &
        13.95 & 10.33 &
        13.51 & 9.66 \\

        Stiff & Smooth &
        11.53 &
        11.50 & 9.96 &
        11.52 & 10.04 &
        11.47 & 10.11 &
        11.61 & \textbf{9.90} \\

        Stiff & Rough &
        10.62 &
        9.72 & \textbf{8.40} &
        9.98 & 8.57 &
        9.85 & 8.56 &
        9.87 & 8.67 \\

        \midrule

        \multicolumn{2}{l}{Overall} &
        10.15 &
        11.02 & \textbf{8.74} &
        11.08 & 8.91 &
        11.09 & 9.48 &
        10.74 & 8.95 \\

        \bottomrule
    \end{tabular}
    }
    \vspace{-3mm}
\end{table}

\subsection{Decision Utility in Executed Manipulation}
\label{sec:results_decision}

We next evaluate physical-context conditioning within the same GPF backbone under executed manipulation (Fig.~\ref{fig:context_utility}(b) and Table~\ref{tab:paired_context}). Although Estimated is more predictive than Hist in simulation, its control gain over GPF-Geom is not detectable because the paired confidence interval includes zero. In contrast, Oracle-concat, Hist-concat, and Hist-FiLM show detectable gains, with Hist-FiLM yielding the lowest observed error. Thus, predictive utility does not directly determine decision
utility (Fig.~\ref{fig:overview}(d)).

Hist-FiLM also reduces error by 0.485\,mm relative to Estimated-concat
(95\% CI $[0.086,\,0.955]$).
The difference between Hist-FiLM and Hist-concat is $-0.158$\,mm
(95\% CI $[-0.680,0.327]$), providing no detectable evidence that FiLM outperforms concatenation.

Crucially, \emph{context utility} also depends on the policy backbone, as shown in Fig.~\ref{fig:context_utility}(c): conditioning produces only small, mixed-sign changes for DLO-DP, whereas the corresponding GPF variants show larger error reductions.

\begin{table}[!t]
    \centering
    \caption{\centering
    Physical-context conditioning on 470 paired simulation AGX trials.
    Gain is relative to GPF-Geom (positive is better); CIs use paired percentile bootstrap.
    }
    \label{tab:paired_context}
    \scriptsize
    \setlength{\tabcolsep}{2.0pt}
    \renewcommand{\arraystretch}{1.04}
    \resizebox{0.75\columnwidth}{!}{%
    \begin{tabular}{@{}lrrrl@{}}
        \toprule
        Controller
        & Error
        & Gain
        & Gain
        & 95\% CI \\
        & (mm)
        & (mm)
        & (\%)
        & (mm) \\
        \midrule

        GPF-Geom
        & 9.257
        & --
        & --
        & -- \\

        Oracle-concat
        & 8.678
        & 0.580
        & 6.3
        & $[0.056,\,1.127]$ \\

        Estimated-concat
        & 8.821
        & 0.436
        & 4.7
        & $[-0.119,\,1.026]$ \\

        Hist-concat
        & 8.494
        & 0.763
        & 8.2
        & $[0.316,\,1.234]$ \\

        \textbf{Hist-FiLM}
        & \textbf{8.336}
        & \textbf{0.921}
        & \textbf{9.9}
        & $[0.386,\,1.500]$ \\

        \bottomrule
    \end{tabular}
    }
    \vspace{-3mm}
\end{table}

\subsection{Closed-Loop Manipulation}
\label{sec:results_closed_loop}

\subsubsection{Simulation}

Closed-loop trials start from more challenging configurations than the one-step evaluation, with mean initial error increasing from 79.2\,mm to 138\,mm. Fig.~\ref{fig:closed_loop}(a) shows that all three controllers progressively reduce error, with \emph{Hist-FiLM} achieving the lowest final mean.

Within the 10-action budget, \emph{Hist-FiLM} reaches 5\,mm in 36\% of trials, compared with 25\% for \emph{GPF-Geom} and 20\% for \emph{DLO-DP}. At 8\,mm, \emph{GPF-Geom} and \emph{Hist-FiLM} achieve similar attainment rates of 60\% and 58\%, respectively, versus 41\% for DLO-DP; at 10\,mm, the rates are 68\% for \emph{GPF-Geom}, 69\% for \emph{Hist-FiLM}, and 59\% for \emph{DLO-DP}. The advantage of history-derived context is therefore most apparent at the stricter 5\,mm threshold, while the two GPF variants perform similarly at coarser tolerances.

\subsubsection{Real robot}\label{sec:real}

\begin{figure}[!t]
    \centering

    \includegraphics[width=\columnwidth]
    {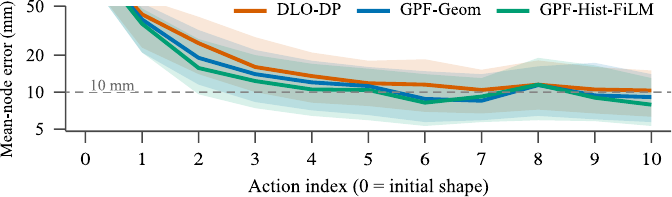}
    \par\vspace{0.2mm}
    {\small (a) Simulation}\par

    \vspace{2.0mm}

    \includegraphics[width=\columnwidth]
    {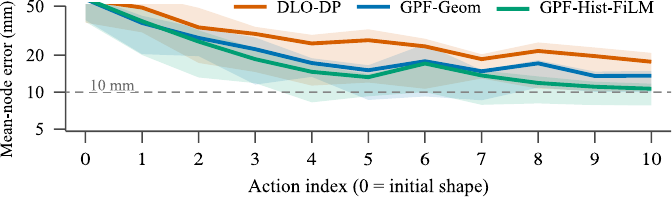}
    \par\vspace{0.2mm}
    {\small (b) Real robot}\par

    \vspace{1mm}

    \caption{
    Closed-loop manipulation in simulation (100 rollouts/controller)
    and on the real robot (60 rollouts/controller).
    Curves show mean-node error and shading shows IQR.
    Panels share a logarithmic scale; action 0 is the initial shape,
    and the dashed line marks 10\,mm.
    }
    \label{fig:closed_loop}
    \vspace{-3mm}
\end{figure}

Fig.~\ref{fig:closed_loop}(b) and Table~\ref{tab:real_multi_step} summarize zero-shot closed-loop performance under the common 10-action budget.
Fig.~\ref{fig:realrobot_process} shows a representative \emph{Hist-FiLM} rollout with the medium DLO on the rough surface, where mean-node error decreases from 107.14\,mm to 7.96\,mm over 10 actions.

\emph{Hist-FiLM} achieves the lowest final mean and rollout-best errors, together with the highest 10\,mm attainment. Its rollout-best error is 1.28\,mm lower than \emph{GPF-Geom} (95\% CI $[0.40,\,2.23]$, paired over 60 target--condition blocks).
Among successful trials, its passage time is comparable to \emph{GPF-Geom}, although this conditional metric should be interpreted jointly with attainment.
Because the simulation and hardware evaluations have different initial-error distributions, Fig.~\ref{fig:closed_loop} supports only within-domain
controller comparisons.

\begin{table}[!t]
    \centering
    \caption{
    Zero-shot real-robot closed-loop performance over 60 rollouts/controller (10 per condition). Best error is mean [median]; passage is mean actions to 10\,mm among successful rollouts. Overall means average conditions; other metrics pool rollouts. Bold marks the best per row.
    }
    \label{tab:real_multi_step}
    \footnotesize
    \setlength{\tabcolsep}{1.5pt}
    \renewcommand{\arraystretch}{1.05}
    \resizebox{\columnwidth}{!}{%
    \begin{tabular}{@{}ll*{9}{c}@{}}
        \toprule
        & &
        \multicolumn{3}{c}{Best error (mm) $\downarrow$}
        & \multicolumn{3}{c}{Reach 10 mm (\%) $\uparrow$}
        & \multicolumn{3}{c}{Passage (steps) $\downarrow$} \\
        \cmidrule(lr){3-5}
        \cmidrule(lr){6-8}
        \cmidrule(lr){9-11}

        DLO & Surface
        & \shortstack{DLO-\\DP} & \shortstack{GPF-\\Geom} & \shortstack{Hist-\\FiLM}
        & \shortstack{DLO-\\DP} & \shortstack{GPF-\\Geom} & \shortstack{Hist-\\FiLM}
        & \shortstack{DLO-\\DP} & \shortstack{GPF-\\Geom} & \shortstack{Hist-\\FiLM} \\
        \midrule

        Soft & Smooth
        & 10.40 [9.82] & 7.97 [7.86] & \textbf{7.85 [7.28]}
        & 50 & \textbf{90} & \textbf{90}
        & 9.2 & 6.7 & \textbf{6.3} \\

        Soft & Rough
        & 10.32 [9.64] & 6.31 [6.62] & \textbf{5.73 [6.16]}
        & 60 & \textbf{100} & \textbf{100}
        & 5.5 & \textbf{5.0} & 5.5 \\

        Medium & Smooth
        & 8.78 [9.69] & 8.36 [8.01] & \textbf{6.73 [7.10]}
        & 50 & 70 & \textbf{100}
        & 8.2 & 5.7 & \textbf{4.6} \\

        Medium & Rough
        & 16.54 [19.37] & 10.48 [10.38] & \textbf{7.42 [7.07]}
        & 20 & 50 & \textbf{90}
        & \textbf{6.0} & 8.0 & 7.2 \\

        Stiff & Smooth
        & 9.31 [9.16] & 7.06 [7.45] & \textbf{6.48 [6.02]}
        & 60 & \textbf{90} & \textbf{90}
        & 6.0 & \textbf{5.1} & 5.7 \\

        Stiff & Rough
        & 11.09 [10.27] & 11.34 [9.82] & \textbf{9.65 [9.87]}
        & 50 & \textbf{60} & \textbf{60}
        & 9.2 & 7.3 & \textbf{7.2} \\

        \midrule

        \multicolumn{2}{l}{Overall}
        & 11.07 [9.79] & 8.59 [8.06] & \textbf{7.31 [7.12]}
        & 48.3 & 76.7 & \textbf{88.3}
        & 7.4 & 6.1 & \textbf{6.0} \\

        \bottomrule
    \end{tabular}
    }%
    \vspace{-4mm}
\end{table}

\section{Discussion}

Physical context can improve response prediction without providing a
corresponding benefit for action selection. In simulation, \emph{Estimated}
context is more predictive than \emph{Hist}, yet yields no detectable one-step
control gain. On hardware, this ordering changes: matched \emph{Hist} gives the
lowest prediction error, while \emph{Hist-FiLM} achieves the strongest observed
closed-loop performance among the deployed controllers. Predictive utility
therefore does not directly imply decision utility.

Joint training alone is unlikely to explain this difference: although the \emph{Hist} encoder is trained jointly with each downstream model while the estimator is frozen, the \emph{Hist} predictor remains less accurate than \emph{Estimated} in simulation. Context
conditioning also affects GPF and the diffusion-policy backbone differently,
suggesting that utility depends both on the information represented and on how
it is integrated into the downstream task. Our experiments do not isolate
these factors causally.

One possible source of this decoupling is that \emph{Hist} can retain
sample-specific response information from recent interactions that is not
captured by the explicit physical parameters. Its strong degradation under
shuffled context supports the importance of such matched information. The
closed-loop results are consistent with this view: in simulation, the benefit
is clearest at the strictest 5\,mm threshold, while \emph{Hist-FiLM} and
\emph{GPF-Geom} perform similarly at 8 and 10\,mm; on hardware,
\emph{Hist-FiLM} also improves rollout-best error and the reported 10\,mm
attainment. This suggests that interaction history may become useful when
geometry alone is insufficient, whether for fine correction or under
sim-to-real mismatch, although our experiments do not isolate these mechanisms.
Passage times in Table~\ref{tab:real_multi_step} are conditioned on success and thus favor controllers that reach 10\,mm only on easier targets, as for \emph{DLO-DP} on the medium--rough condition; they should therefore be read jointly with attainment.

The stronger zero-shot performance of \emph{Hist} on hardware may reflect
task-relevant interaction information beyond the explicit $(E,\mu)$
parameterization, but could also arise from sim-to-real degradation of the
parameter estimator. Our experiments do not determine which additional
properties, if any, \emph{Hist} encodes. Moreover, exact state pairing is not
possible on the real robot because manual reset cannot reproduce the same
initial state; trials are therefore matched only by target and condition, with
one execution per controller--target pair.

\vspace{-2mm}
\section{Conclusion}

In this paper, we investigated whether physical context that improves prediction of DLO response also improves executed manipulation.
Our results show that \emph{predictive utility} does not necessarily imply \emph{decision utility}: the representations that best improve prediction are not those that most benefit action selection.

Among the representations considered, interaction-history context provides the strongest observed control performance. Its benefit is clearest at the strictest threshold in simulation and persists under zero-shot transfer to physical DLO manipulation, without real-world policy training or fine-tuning.
Overall, these results suggest that physical context should be evaluated according to the decisions it
supports, rather than predictive accuracy alone.

\balance
\bibliographystyle{IEEEtran}
\bibliography{Reference}
\end{document}